%% file: paper.tex
\documentclass[sigconf]{acmart}
\usepackage{microtype} % 在导言区加入这行，通常能解决 90% 的溢出问题
\usepackage{multirow}
\usepackage{algorithm}
\usepackage{algorithmic}

\AtBeginDocument{%
  }

\acmConference[EASE 2026]{The 30th International Conference on Evaluation and Assessment in Software Engineering}{9–12 June, 2026}{Glasgow, Scotland, United Kingdom}
  
\begin{document}
% \begin{sloppypar}

% title, author
\input{sections/0-header.tex}
% abstract, CSSXML, keywords, received info
\input{sections/1-abstract.tex}

\maketitle

% contents begin here
\input{sections/2-intro.tex}
\input{sections/4-methodology.tex}

\input{sections/5-experiments.tex}
\input{sections/6-related-work.tex}
\input{sections/7-conclusion.tex}
\input{sections/8-ack.tex}
% contents end here

%%
%% Print the bibliography
%%

\bibliographystyle{ACM-Reference-Format}
\bibliography{references}

%%
%% If your work has an appendix, this is the place to put it.
% \appendix

% \end{sloppypar}
\end{document}

%% file: sections/0-header.tex
%%
%% The "title" command has an optional parameter,
%% allowing the author to define a "short title" to be used in page headers.
\title{STLGT: A Scalable Trace-Based Linear Graph Transformer for Tail Latency Prediction in Microservices}

%%
%% The "author" command and its associated commands are used to define
%% the authors and their affiliations.
%% Of note is the shared affiliation of the first two authors, and the
%% "authornote" and "authornotemark" commands
%% used to denote shared contribution to the research.

\author{Yongliang Ding}
\affiliation{%
  \institution{East China Normal University}
  \city{Shanghai}
  \country{China}}
\email{ylding@stu.ecnu.edu.cn}

\author{Qigong Bi}
\affiliation{%
  \institution{East China Normal University}
  \city{Shanghai}
  \country{China}}
\email{qgbi@stu.ecnu.edu.cn}

% \author{Yun Yang}
% \affiliation{%
%   \institution{East China Normal University}
%   \city{Shanghai}
%   \country{China}}
% \email{yyang@stu.ecnu.edu.cn}

\author{Peng Pu}
\affiliation{%
  \institution{East China Normal University}
  \city{Shanghai}
  \country{China}}
\email{ppu@cc.ecnu.edu.cn}

%%
%% By default, the full list of authors will be used in the page
%% headers. Often, this list is too long, and will overlap
%% other information printed in the page headers. This command allows
%% the author to define a more concise list
%% of authors' names for this purpose.
\renewcommand{\shortauthors}{Yongliang Ding et al.}

%% file: sections/1-abstract.tex
%%
%% The abstract is a short summary of the work to be presented in the
%% article.
\begin{abstract}
  Accurate end-to-end tail-latency forecasting is critical for proactive SLO management in microservice systems.
  However, modeling long-range dependency propagation and non-stationary, bursty workloads while maintaining inference efficiency at scale remains challenging.
  We present STLGT (\textbf{S}calable \textbf{T}race-based \textbf{L}inear \textbf{G}raph \textbf{T}ransformer), a per-API predictor that encodes traces as span graphs for multi-step p95 tail-latency forecasting.
  STLGT uses a \emph{structure-aware} linear graph Transformer to propagate cross-service dependencies with inference time linear in span graph size, and a decoupled temporal module to capture workload dynamics.
  Across a personalized education microservice application and open-source microservice benchmarks~\cite{deathstarbench, ali-traces}, STLGT improves forecasting accuracy over PERT-GNN~\cite{pert} by 8.5\% MAPE on average and achieves up to 12$\times$ faster CPU inference at $N=32$, matching the maximum span graph size after preprocessing the Alibaba traces~\cite{ali-traces}. Ablation studies further demonstrate the effectiveness of each component, especially under bursty traffic.
\end{abstract}

%%
%% The code below is generated by the tool at http://dl.acm.org/ccs.cfm.
%% Please copy and paste the code instead of the example below.
%%

\ccsdesc[500]{Computing methodologies~Neural networks}
\ccsdesc[500]{Computing methodologies~Supervised learning by regression}

%%
%% Keywords. The author(s) should pick words that accurately describe
%% the work being presented. Separate the keywords with commas.
\keywords{microservices, tail latency prediction, proactive auto-scaling, distributed tracing, graph Transformer}

% \received{20 February 2007}
% \received[revised]{12 March 2009}
% \received[accepted]{5 June 2009}

%% file: sections/2-intro.tex
\section{Introduction}
\label{sec:introduction}

Cloud-native computing has become a foundational paradigm for modern software delivery, with microservices and container technologies serving as its core architectural enablers \cite{surveyautoscaling2018chen, serveymicroservices2017dragoni, serveycloud2024deng}. Consequently, major cloud providers such as Microsoft, IBM, and Alibaba have adopted Kubernetes-based container orchestration platforms to manage large-scale production systems \cite{autopilot2020rzadca, orion2022mahgoub, aware, firm, harmonybatch2024chen, ahpa2023zhang, luo2024stamp}.
This trend is also visible in digital education platforms, where learning, assessment, recommendation, content delivery, and analytics services are increasingly decomposed into independently deployed microservices.
Unlike many steady enterprise workloads, education workloads often contain schedule-driven bursts: online exams, deadline-driven submissions, and synchronous classroom activities can trigger sharp traffic increases within a short time window.
Such bursts are difficult to absorb reactively because latency degradation may propagate from assessment APIs to authentication, question-bank, grading, and analytics services, making accurate short-term tail-latency prediction particularly important.

% %% v1
% Elasticity is a key characteristic of cloud services \cite{deepscaler}. Compared with monolithic architectures, microservice-based systems typically comprise hundreds or thousands of service instances deployed across homogeneous or heterogeneous clusters, with virtualization and containerization abstracting underlying hardware heterogeneity. By continuously monitoring runtime metrics such as CPU and memory utilization, resources can be dynamically provisioned or reclaimed to meet Quality of Service (QoS) requirements while minimizing operational costs, a process commonly referred to as scaling.

% However, the rapid growth in the number of microservices significantly increases the cost of manual intervention, rendering it impractical in large-scale systems \cite{surveydevops2019leite}. As a result, auto-scaling has become a critical capability in modern microservice architectures.

%% v2
Elasticity is a key characteristic of cloud services \cite{deepscaler}. Compared with monolithic architectures, microservice-based systems typically comprise hundreds or thousands of service instances deployed across homogeneous or heterogeneous clusters, with virtualization and containerization abstracting underlying hardware heterogeneity. Elasticity is realized through scaling: resources are provisioned or reclaimed based on runtime metrics (e.g., CPU and memory utilization) to meet QoS requirements while controlling operational costs. As the number of microservices grows rapidly, manual scaling becomes increasingly costly and impractical in lar\-ge-scale systems \cite{surveydevops2019leite}, making auto-scaling a critical capability in modern microservice architectures.

%% v2
Existing auto-scaling approaches generally fall into two categories: reactive and proactive \cite{pert}. Reactive approaches are typically triggered by predefined thresholds or rules and are simple to implement; however, their delayed response to dynamic workload variations often leads to Service Level Objective (SLO) violations. In contrast, proactive approaches aim to anticipate future workload demands or QoS states and provision resources in advance, thereby reducing the risk of SLO violations. As a result, proactive methods have become an important direction in auto-scaling and cloud resource management research.

A typical proactive auto-scaling system consists of two core components: a predictor and a scaling controller. While the design of scaling controllers has been extensively studied—including reinforcement-learning-based controllers (e.g., actor-critic methods) for adaptive resource allocation \cite{aware,gwydion}—the modeling capability of predictors remains relatively limited, particularly in complex microservice environments. We highlight three key challen\-ges:

%% v2
\begin{figure}[ht]
  \centering
  \includegraphics[width=\linewidth]{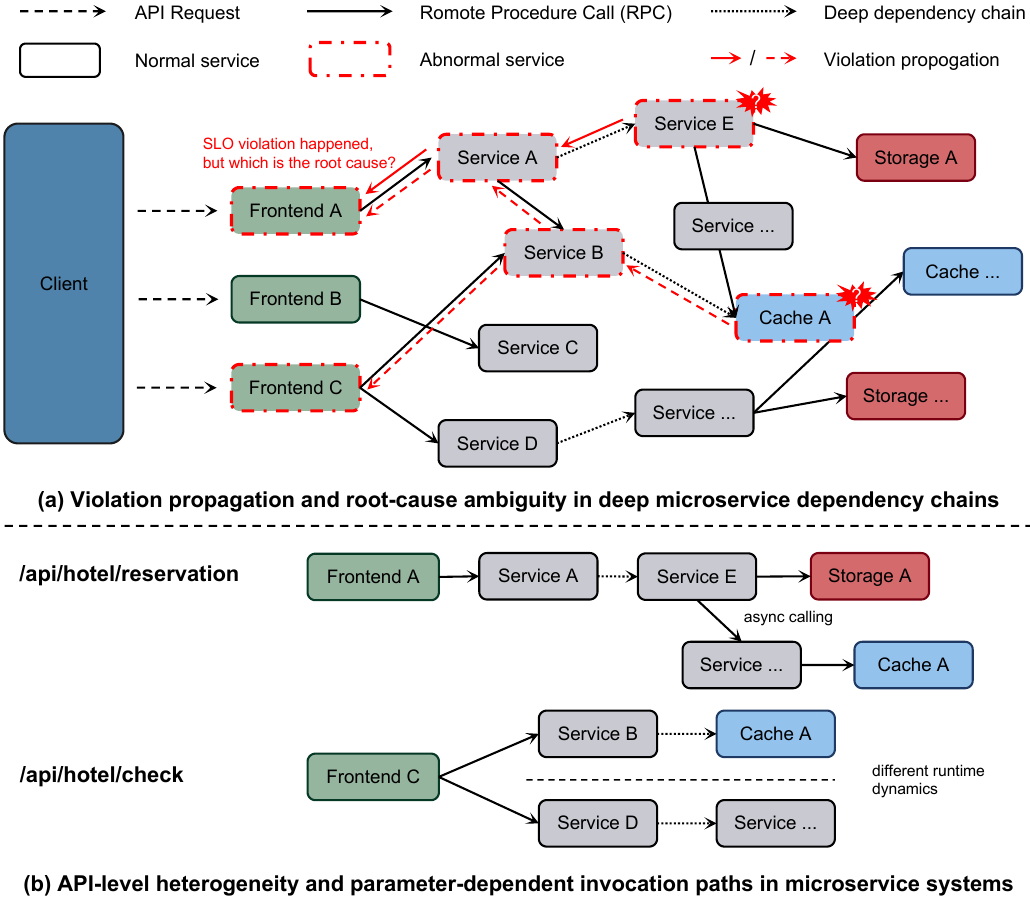}
  \caption{Challenges in Modeling Long-Chain Dependencies and Runtime Heterogeneity in Microservice Call Graphs}
  \Description{
    This figure illustrates key challenges of existing microservice call graph (MCG) modeling approaches. 
    (a) In deep dependency chains, performance degradations or failures in downstream services can propagate and compound along invocation paths, leading to accumulated prediction errors and ambiguous root causes ambiguity at upstream services when only coarse-grained or local dependencies are considered. 
    (b) Real-world API invocations exhibit significant runtime heterogeneity, where different APIs, asynchronous execution, and parameter-dependent requests induce distinct and dynamically changing invocation paths, posing substantial challenges to accurate and scalable dependency modeling.
  }
  \label{fig:mcg}
\end{figure}

\begin{enumerate}
    \item \textbf{C1: Limited global dependency modeling.} Many existing predictors either ignore microservice invocation dependencies or capture only local neighborhood information, ma\-king them insufficient to capture global dependen\-cy propagation across invocation paths. As a result, prediction errors can accumulate along long dependency chains, which may trigger mis-scaling decisions and ultimately increase the risk of SLO violations in large-scale microservice systems (Fig.~\ref{fig:mcg}(a)).

    \item \textbf{C2: Insufficient burst-aware temporal modeling.} Many existing approaches implicitly assume workload stationarity or periodicity and rely on short historical windows, limiting their ability to capture non-periodic and bursty traffic patterns commonly observed in production environments, especially under traffic hotspots or sudden demand surges.

    \item \textbf{C3: Poor scalability and limited faithfulness to production characteristics.} Existing methods often adopt ove\-rly simplified abstractions of service APIs, ignoring heterogeneity in request rates, parameter-dependent invocation paths, and large number of APIs. Moreover, predictors based on graph neural networks or global attention mechanisms typically incur substantial computational overhead as system scale grows, significantly constraining their applicability in large-scale deployments (Fig.\ref{fig:mcg}(b)).
\end{enumerate}

To address these challenges, we propose \textbf{STLGT}: a \underline{\textbf{S}}calable \underline{\textbf{T}}r\-ace-based \underline{\textbf{L}}inear \underline{\textbf{G}}raph \underline{\textbf{T}}ransformer for proactive microservice tail-latency prediction.
STLGT is a scalable, trace-based \emph{per-API} predictor: for each API trace stream, it constructs an API-specific span graph from distributed traces and employs a \emph{structure-aware} linear graph Transformer encoder to capture long-range dependency propagation and cross-path influences.
This trace-driven partitioning bounds the feature-graph size per prediction instance and avoids graph size blow-up.
To model non-stationary and bursty workload dynamics without incurring the overhead of coupled spatiotemporal attention, STLGT decouples global dependency encoding from temporal dynamics modeling via a dedicated temporal module.
We assume the dominant call path of each API is stable within an evaluation period; when the call path changes, the span graph can be refreshed from recent traces.

Our contributions are summarized as follows:

\begin{itemize}
    \item \textbf{Trace-based span graph abstraction for tail-latency pr\-ediction} (\emph{solves C1 and C3}).
    We introduce an API-level span graph representation and a lightweight span-to-graph construction derived from distributed traces, enabling efficient feature alignment for tail-latency prediction.

    \item \textbf{Structure-aware linear graph Transformer for global dependency propagation} (\emph{solves C1 and C3}).
    We design a structure-aware linear graph Transformer encoder whose inference cost scales linearly with graph size while capturing global dependency propagation for tail-latency prediction.

    \item \textbf{Decoupled spatiotemporal modeling for scalable inference} (\emph{solves C2 and C3}).
    We decouple global dependency encoding from temporal dynamics modeling through a dedicated temporal module, avoiding expensive coupled spatiotemporal attention and reducing inference complexity to an additive linear form.
\end{itemize}

The remainder of this paper is organized as follows.
Section~\ref{sec:problem-definition} defines the problem.
Sections~\ref{sec:data-collector} and~\ref{sec:latency-predictor} present the methodology.
Section~\ref{sec:experiments} reports experimental results.
Section~\ref{sec:related-work} reviews related work, and Section~\ref{sec:conclusion} concludes the paper.

%% file: sections/4-methodology.tex
\begin{figure*}[t]
  \centering
  \includegraphics[width=\linewidth]{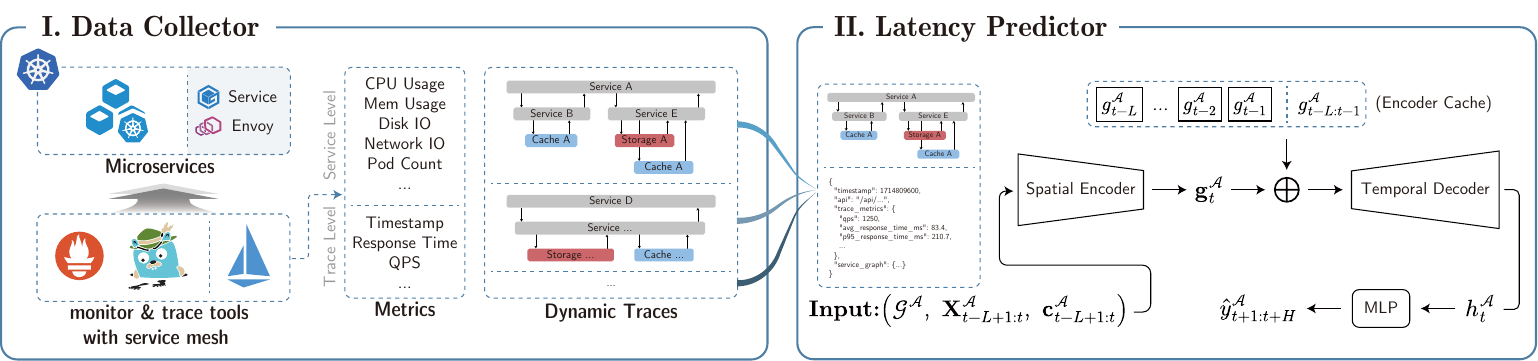}
  \caption{The overall architecture of STLGT}
  \Description{
    The figure illustrates the overall architecture of STLGT.
    At each time step, a data collector aggregates node-level metrics and distributed traces, and constructs a directed span graph with time-aligned features.
    A linear graph Transformer encoder jointly encodes the graph structure and the multi-source features to produce node representations.
    A trace-aware readout layer aggregates node representations into a fixed-dimensional trace embedding, which is combined with historical embeddings and fed into a lightweight temporal decoder.
    The prediction head outputs $H$-step-ahead tail latency at the target quantile for proactive decision making.
  }
  \label{fig:stlgt}
\end{figure*}

\input{sections/3-problem-definition.tex}

% \section{Methodology}
% \label{sec:methodology}

% \input{sections/method/4-1-definition.tex}
\input{sections/method/4-2-data-collector.tex}
\input{sections/method/4-3-predictor.tex}

%% file: sections/3-problem-definition.tex
\section{Problem Definition}
\label{sec:problem-definition}

This section formalizes the trace-based microservice p95 tail-laten\-cy prediction problem studied in this paper.
We define the microservice call graph and distributed tracing primitives (\emph{trace} and \emph{span}), then define how we induce an API-specific span graph.
Since raw traces must be aggregated into window-aligned learning signals in any practical pipeline, we also introduce concise aggregation operators that map traces to the tail-latency label and to node features.
Our definitions follow standard distributed tracing terminology~\cite{davidson2023tracing} and prior trace-graph modeling work~\cite{pert,fast-pert}.

\noindent\textbf{Definition 1: Microservice Call Graph.}
For a target API endpoint $\mathcal{A}$, we denote $\mathcal{G}=(\mathcal{V},\mathcal{E})$ as a directed \emph{Microservice Call Graph (MCG)}, where
$\mathcal{V}=\{s_{1}, s_{2}, \dots, s_{M}\}$ is the set of $M$ microservices and
$\mathcal{E}\subseteq \mathcal{V}\times\mathcal{V}$ is the set of directed invocation relations.
An edge $(s_{i}, s_{j})\in \mathcal{E}$ indicates that microservice $s_i$ may invoke microservice $s_j$ via RPC.

\noindent\textbf{Definition 2: Trace and Span.}
We discretize time into windows of fixed length $\Delta$ and use $t$ to index windows.
For a target API endpoint $\mathcal{A}$, let $\mathcal{T}^{\mathcal{A}}_{t}$ denote the set of traces of $\mathcal{A}$ observed within window $t$.
Each trace $\tau \in \mathcal{T}^{\mathcal{A}}_{t}$ is uniquely identified by \texttt{traceID}$(\tau)$ and consists of a set of spans
$\Sigma(\tau)=\{\sigma_{\tau,1}, \sigma_{\tau,2}, \dots, \sigma_{\tau,|\tau|}\}$,
where all spans in $\Sigma(\tau)$ share the same \texttt{traceID}$(\tau)$ but have distinct \texttt{spanID}s.
A span $\sigma$ captures one invocation instance in the trace: it indicates that a microservice $s\in \mathcal{V}$ is invoked by a caller microservice $p\in \mathcal{V}$ during a time interval $[t_{1}, t_{2}]$:
\begin{equation}
  \sigma = \big(\texttt{traceID},\ \texttt{spanID},\ p,\ s,\ t_{1},\ t_{2}\big).
\end{equation}
Spans are connected via parent--child relations, forming a rooted tree that captures causal execution dependencies within a trace.

\noindent\textbf{Definition 3: API-Induced Span Graph.}
Given $\mathcal{G}=(\mathcal{V},\mathcal{E})$ and a target API $\mathcal{A}$, we induce an API-specific \emph{Span Graph} $\mathcal{G}^{\mathcal{A}}=(\mathcal{V}^{\mathcal{A}},\\\mathcal{E}^{\mathcal{A}})$, where $\mathcal{V}^{\mathcal{A}}=\{v^{\mathcal{A}}_{1}, v^{\mathcal{A}}_{2}, \dots, v^{\mathcal{A}}_{N}\}$ is a set of $N$ execution-stage nodes and $\mathcal{E}^{\mathcal{A}}$ encodes their invocation dependencies.
Each node $v^{\mathcal{A}}_{i}\in \mathcal{V}^{\mathcal{A}}$ corresponds to a span graph representation of some microservice $s\in \mathcal{V}$.
We define a surjective mapping $\pi:\mathcal{V}^{\mathcal{A}}\rightarrow \mathcal{V}$ that maps each stage node to its corresponding microservice.
Since a microservice may be invoked multiple times within a trace, multiple nodes in $\mathcal{V}^{\mathcal{A}}$ may map to the same microservice, and therefore $N \ge M$.
Each directed edge $(v^{\mathcal{A}}_{i}, v^{\mathcal{A}}_{j})\in \mathcal{E}^{\mathcal{A}}$ indicates that the stage represented by $v^{\mathcal{A}}_{i}$ is a parent of $v^{\mathcal{A}}_{j}$ under the trace-defined span hierarchy.
By construction, each span graph edge also corresponds to a valid microservice-level invocation, i.e., $(\pi(v^{\mathcal{A}}_{i}),\pi(v^{\mathcal{A}}_{j}\\))\in \mathcal{E}$.
In this paper, we treat both $\mathcal{G}$ and $\mathcal{G}^{\mathcal{A}}$ as \textbf{time-invariant} structures, and use the time-dependent trace set $\mathcal{T}^{\mathcal{A}}_{t}$ to capture runtime dynamics in each window.

\noindent\textbf{Definition 4: Trace Aggregation and Feature Construction.}
We introduce window-level aggregation operators that map raw traces to learning signals.
First, the tail-latency label in window $t$ is defined as
\begin{equation}
  y^{\mathcal{A}}_{t} = \rho_{q}\big(\mathcal{T}^{\mathcal{A}}_{t}\big),
\end{equation}
where $\rho_{q}(\cdot)$ returns the $q$-quantile of end-to-end trace latencies in $\mathcal{T}^{\mathcal{A}}_{t}$.
For simplicity, we fix $q=0.95$ and predict p95 tail latency in this paper.
Second, we extract a trace-common vector $\mathbf{c}^{\mathcal{A}}_{t}\in\mathbb{R}^{d_c}$ from $\mathcal{T}^{\mathcal{A}}_{t}$,
\begin{equation}
  \mathbf{c}^{\mathcal{A}}_{t} = \phi_{\text{common}}\big(\mathcal{T}^{\mathcal{A}}_{t}\big),
\end{equation}
which summarizes API-level workload and performance signals such as throughput, response-time percentiles (p50/p90/p99), and failure ratio.
Third, for each stage node $v^{\mathcal{A}}_{i}$, we extract a span-specific vector $\mathbf{u}^{\mathcal{A}}_{t,i}\in\mathbb{R}^{d_u}$ by aggregating spans in window $t$ that correspond to this stage:
\begin{equation}
  \mathbf{u}^{\mathcal{A}}_{t,i} = \phi_{\text{span}}\big(\mathcal{T}^{\mathcal{A}}_{t},\ v^{\mathcal{A}}_{i}\big),
\end{equation}
which captures fine-grained timing structure within traces, including normalized start/end time and normalized duration.
Finally, let $\mathbf{m}^{s}_{t}\in\mathbb{R}^{d_s}$ denote the service-level metrics of microservice $s$ collected in window $t$.
The input feature of stage node $v^{\mathcal{A}}_{i}$ is constructed as
\begin{equation}
  \mathbf{x}^{\mathcal{A}}_{t,i}
  =
  \Big[
    \mathbf{m}^{\pi(v^{\mathcal{A}}_{i})}_{t}\ \Vert\ \mathbf{u}^{\mathcal{A}}_{t,i}
  \Big],
\end{equation}
Then we denote $\mathbf{X}^{\mathcal{A}}_{t}=[\mathbf{x}^{\mathcal{A}}_{t,1};\ \dots;\ \mathbf{x}^{\mathcal{A}}_{t,N}]$ as the node-feature matrix for $\mathcal{A}$ in window $t$.

\noindent\textbf{Problem: Tail-Latency Prediction.}
Given a history length $L$ and a prediction horizon $H$, the goal is to learn a predictor $f_{\theta}$ that predicts the next-$H$ tail latencies from the past-$L$ windows:
\begin{equation}
  \hat{\mathbf{y}}^{\mathcal{A}}_{t+1:t+H}
  = f_{\theta}\Big(\mathcal{G}^{\mathcal{A}},\ \mathbf{X}^{\mathcal{A}}_{t-L+1:t},\ \mathbf{c}^{\mathcal{A}}_{t-L+1:t} \Big)
  \in \mathbb{R}^{H}.
\label{eq:prediction}
\end{equation}
Here $\mathbf{X}^{\mathcal{A}}_{t-L+1:t}$ denotes the feature sequence $\{\mathbf{X}^{\mathcal{A}}_{t-L+1},\ \dots\ ,\ \mathbf{X}^{\mathcal{A}}_{t}\}$ constructed from traces and service-level metrics.

%% file: sections/method/4-2-data-collector.tex
\section{Data Collector}
\label{sec:data-collector}

In microservice systems, a common and practical way to obtain fine-grained RPC invocation trajectories is through distributed tra\-cing \cite{davidson2023tracing}. By instrumenting each incoming request and recording its execution as a collection of causally related spans, tracing systems capture the complete end-to-end execution path across service boundaries, naturally exposing the parent–child relationships and temporal dependencies among RPC calls. In modern cloud-native deployments, service meshes such as Istio \cite{istio} provide transparent tracing support by injecting instrumentation into the data plane via sidecar proxies, enabling request-level observability without requiring application-level code changes. The resulting trace data are typically collected and managed by dedicated backends such as Jaeger \cite{jaeger}, which preserve span hierarchies and timing information for downstream analysis. As demonstrated in prior works, tracing-based RPC trajectories constitute a reliable and semantically rich foundation for constructing span graphs and reasoning about latency propagation along execution paths.

\subsection{Runtime Metrics Definition}

\begin{table}[t]
  \centering
  \caption{Runtime metrics collected from monitoring and tracing systems.}
  \label{tab:metrics}
  \vspace{2pt}
  \begin{tabular}{l|c|l}
    \toprule
    \textbf{Level} & \textbf{Subtype} & \textbf{Metric} \\
    \midrule
    \multirow{6}{*}{Service Level}
      & \multirow{6}{*}{--} & pod\_count \\
      &  & cpu\_usage\_ratio \\
      &  & memory\_usage\_ratio \\
      &  & network\_rx\_bytes \\
      &  & network\_tx\_bytes \\
      &  & disk\_io\_bytes \\
    \midrule
    \multirow{10}{*}{Trace Level}
      & \multirow{7}{*}{\emph{trace-common}} & api\_throughput \\
      &  & response\_latency\_p50 \\
      &  & response\_latency\_p90 \\
      &  & response\_latency\_p99 \\
      &  & response\_latency\_avg \\
      &  & response\_latency\_median \\
      &  & failure\_ratio \\
      \cmidrule(lr){2-3}
      & \multirow{3}{*}{\emph{span-specific}} & span\_start\_time\_norm \\
      &  & span\_end\_time\_norm \\
      &  & span\_duration\_norm \\
    \bottomrule
  \end{tabular}
\end{table}

For a microservice cluster, we collect a set of runtime metrics summarized in Table~\ref{tab:metrics}, covering both service-level and trace-level observations.
These metrics are designed to jointly capture the resource provisioning state of the cluster and the execution characteristics of individual API requests, which together form the basis for subsequent end-to-end latency analysis and modeling.

\emph{Service-level metrics} describe the coarse-grained operating conditions of individual microservices, independent of specific request executions.
They capture resource availability and load conditions from a resource-centric perspective, including service scaling status and resource utilization.
Such metrics indicate whether a microservice is under-provisioned, resource-constrained, or lightly loaded, and therefore provide essential contextual information for interpreting variations in request-level latency.
As summarized in Table~\ref{tab:metrics}, we focus on representative indicators of computation, me\-mory pressure, and network activity, which are known to be primary factors influencing microservice latency \cite{sinan,deepscaler,wang2024deepscaling,pbscaler}.

\emph{Trace-level metrics} characterize the performance of individual API invocation traces and are inherently request-centric.
Given a trace $\tau$, its trace-level representation lies in $\mathbb{R}^{d_\tau}$ and can be decomposed as
$\mathbb{R}^{d_{\tau\text{-common}}} \oplus \mathbb{R}^{d_{\tau\text{-specific}}}$.
The \emph{trace-common} component captures API-aware performance indicators, such as request throughput, response-time latency, and failure ratio.
These metrics enable fine-grained QoS modeling across different APIs, as distinct APIs often exhibit heterogeneous execution costs and latency expectations. For example, CPU-intensive APIs often exhibit higher tail latency than lightweight request handlers.
In contrast, the \emph{span-specific} component encodes fine-grained temporal information wi\-thin a trace, including per-span relative start times and execution durations normalized to the trace lifetime.
Such metrics reflect the synchronous and asynchronous invocation patterns among micros\-ervices, and preserve the internal execution structure of RPC paths \cite{pert}.
These span-level signals will be further exploited when constructing the span graphs introduced in the following section.

\subsection{Span Graph Construction from Traces}

For a trace $\tau$, we denote its set of spans as $\Sigma(\tau)$, where each span record contains the caller microservice ($p$), the callee microservice ($s$), and timing information (start/end timestamps and duration).

\begin{algorithm}[t]
  \caption{Span Graph Construction and Feature Aggregation}
  \label{alg:span-graph-construction}
  \begin{algorithmic}[1]
  \REQUIRE MCG $\mathcal{G}=(\mathcal{V},\mathcal{E})$; service metrics $\{\mathbf{m}^s_t\}_{s\in\mathcal{V}}$; traces of API $\mathcal{A}$ in window $t$: $\mathcal{T}^{\mathcal{A}}_{t}$.
  \ENSURE Span graph $\mathcal{G}^{\mathcal{A}}=(\mathcal{V}^{\mathcal{A}}, \mathcal{E}^{\mathcal{A}})$; node-feature matrix $\mathbf{X}^{\mathcal{A}}_{t}$; trace-common context $\mathbf{c}^{\mathcal{A}}_{t}$.

  %% step 1. initialization.
  \STATE $\mathcal{V}^{\mathcal{A}}, \mathcal{E}^{\mathcal{A}}  \leftarrow \emptyset, \emptyset$
  \STATE $\texttt{vid} \leftarrow \emptyset$ \quad $\triangleright$ map: stage node $v$ $\mapsto$ row id in $\mathbf{X}^{\mathcal{A}}_{t}$
  \STATE $\mathbf{c}^{\mathcal{A}}_{t} \leftarrow \phi_{\text{common}}(\mathcal{T}^{\mathcal{A}}_{t})$

  %% step 2. node discovery.
  \FORALL{$\tau \in \mathcal{T}^{\mathcal{A}}_{t}$}
      \FORALL{$\sigma \in \Sigma(\tau)$}
          \STATE $v \leftarrow (\sigma.\mathsf{parent},\ \sigma.\mathsf{s})$
          \IF{$v \notin \texttt{vid}$}
              \STATE $\mathcal{V}^{\mathcal{A}} \leftarrow \mathcal{V}^{\mathcal{A}} \cup \{v\}$
              \STATE $\texttt{vid}[v] \leftarrow |\mathcal{V}^{\mathcal{A}}|$
          \ENDIF
      \ENDFOR
  \ENDFOR

  %% step 3. edge construction.
  \STATE $\texttt{out} \leftarrow \emptyset$ \quad $\triangleright$ map: caller $p$ $\mapsto$ stage nodes $(p,\cdot)$
  \FORALL{$v=(p,s) \in \mathcal{V}^{\mathcal{A}}$}
      \STATE $\texttt{out}[p].\mathsf{append}(v)$
  \ENDFOR
  \FORALL{$v=(p,s) \in \mathcal{V}^{\mathcal{A}}$}
      \FORALL{$v' \in \texttt{out}[s]$}
          \STATE $\mathcal{E}^{\mathcal{A}} \leftarrow \mathcal{E}^{\mathcal{A}} \cup \{(v,v'), (v',v)\}$ \quad $\triangleright$ add reverse edges for bidirectional message passing
      \ENDFOR
  \ENDFOR

  %% step 4. feature construction.
  \STATE $\mathbf{X}^{\mathcal{A}}_{t} \leftarrow \emptyset$
  \FORALL{$v=(p,s) \in \mathcal{V}^{\mathcal{A}}$}
      \STATE $\mathbf{u}^{\mathcal{A}}_{t}(v) \leftarrow \phi_{\text{span}}(\mathcal{T}^{\mathcal{A}}_{t}, v)$
      \STATE $i \leftarrow \texttt{vid}[v]$
      \STATE $\mathbf{X}^{\mathcal{A}}_{t}[i] \leftarrow [\mathbf{m}^{s}_{t}\ \Vert\ \mathbf{u}^{\mathcal{A}}_{t}(v)]$
  \ENDFOR

  \STATE \textbf{return} $\mathcal{G}^{\mathcal{A}}=(\mathcal{V}^{\mathcal{A}}, \mathcal{E}^{\mathcal{A}})$, $\mathbf{X}^{\mathcal{A}}_{t}$, $\mathbf{c}^{\mathcal{A}}_{t}$
  \end{algorithmic}
\end{algorithm}

We construct an API-specific \emph{Span Graph} $\mathcal{G}^{\mathcal{A}}=(\mathcal{V}^{\mathcal{A}},\mathcal{E}^{\mathcal{A}})$ from distributed traces, as summarized in Algorithm~\ref{alg:span-graph-construction}. 
This design is inspired by PERT-style trace graphs~\cite{pert,fast-pert} in that it leverages per-request span hierarchies to expose cross-service execution structure. 
However, unlike PERT/FastPERT which may split each span into multiple stages to explicitly encode request/response ordering and sequential versus parallel calls, we keep a compact node set keyed by trace-defined caller--callee stages and attach normalized start time, normalized end time, and normalized duration as span-level features.
This avoids graph size blow-up and reduces construction overhead, while preserving the execution evidence recorded by tracing and enabling efficient feature alignment for downstream latency prediction.

Concretely, given a MCG $\mathcal{G}=(\mathcal{V},\mathcal{E})$ and traces $\mathcal{T}^{\mathcal{A}}_{t}$ observed in window $t$, we index each execution-stage node by a pair $v=(p,s)$ extracted from spans $\sigma\in\Sigma(\tau)$, where $p=\sigma.\mathsf{parent}$ and $s=\sigma.\mathsf{s}$. Each unique $v$ is added to $\mathcal{V}^{\mathcal{A}}$ at its first occurrence. To preserve caller-specific contexts, distinct nodes ${(p_1,s),(p_2,s),\ldots}$ are maintained even for the same callee $s$.
To implement the span-specific aggregation $\phi_{\text{span}}(\mathcal{T}^{\mathcal{A}}_{t},v)$, we first aggregate repeated spans that map to the same stage within each trace by taking the earliest normalized start time, the latest normalized end time, and the summed normalized duration, then average these per-trace statistics across traces in the window. This produces a stable per-request stage signal and naturally normalizes the aggregated duration by the API throughput in the window.

For edge construction, we connect caller--callee stages along the invocation structure. For any stage node $v=(p,s)$, we add edges from $v$ to all stages $v'=(s,u)$ that are invoked by $s$ in the same API span graph, and we also add reverse edges to enable bidirectional information flow during graph encoding.

%% file: sections/method/4-3-predictor.tex
\section{Latency Predictor}
\label{sec:latency-predictor}

Given the API-specific span graph $\mathcal{G}^{\mathcal{A}}=(\mathcal{V}^{\mathcal{A}},\mathcal{E}^{\mathcal{A}})$, time-varying node features $\mathbf{X}_{t-L+1:t}$ and trace-common contexts $\mathbf{c}_{t-L+1:t}$ constructed in Sec.~\ref{sec:data-collector},
STLGT factorizes the predictor $f_\theta$ (Eq.~\ref{eq:prediction}) into three stages:
(i) a \emph{linear} graph Transformer encoder that produces node representations per time window,
(ii) a readout module that aggregates node representations and fuses trace-common context into a fixed-dimensional window embedding,
and (iii) a lightweight temporal decoder that predicts the next-$H$ tail latencies from the last-$L$ embeddings.

% 感觉没必要写这段
% Throughout this section, we use lightweight MLPs to implement point-wise learnable mappings applied row-wise to node or time-step representations.
% Specifically, unless otherwise noted, a one-layer MLP takes the form $\mathrm{MLP}(\mathbf{X})=\sigma(\mathbf{X}\mathbf{W}+\mathbf{b})$, where $\sigma(\cdot)$ is a point-wise activation.
% When $\sigma(\cdot)$ is set to the identity, the one-layer MLP reduces to an affine (linear) layer.
% Different MLP blocks use independent parameters, and we distinguish them by subscripts (e.g., $\mathrm{MLP}_{\text{in}}$, $\mathrm{MLP}_{\text{ts}}$, and $\mathrm{MLP}_{\text{out}}$).

\subsection{Spatial Encoder}

\begin{figure*}[t]
  \centering
  \includegraphics[width=\linewidth]{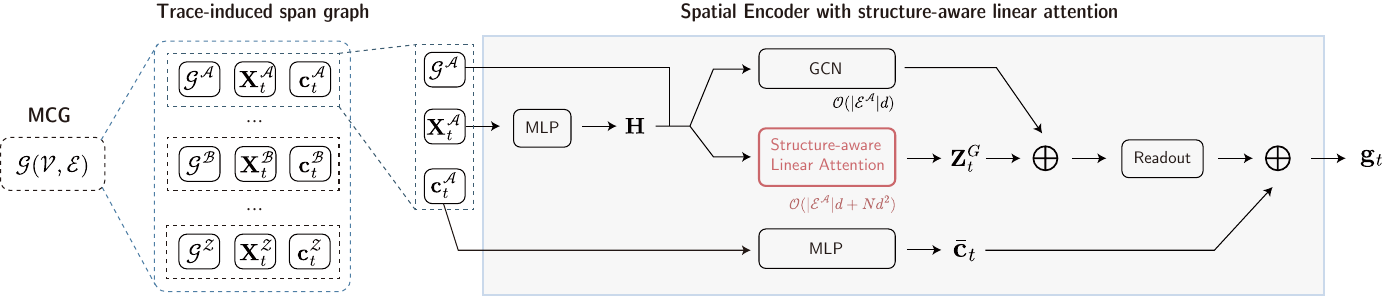}
  \caption{Spatial encoder architecture of STLGT with the Structure-aware linear attention module.}
  \Description{
    The figure illustrates the spatial encoder of STLGT, where the highlighted Structure-aware module injects the span graph topology into global mixing.
    Node features are first mapped to the hidden space, then processed by a structure-aware global mixing branch (linear attention with one-shot sparse pre-mixing) and a local propagation branch (GCN) in parallel.
    The two branch outputs are fused by a feed-forward layer with a residual connection and layer normalization to produce the final node representations.
  }
  \label{fig:encoder}
\end{figure*}

%% step 1. 非线性投影，将 d_in 投影到 d 维度的超空间中
Figure~\ref{fig:encoder} shows the spatial encoder pipeline, including our structu\-re-aware module.
Let $N=|\mathcal{V}^{\mathcal{A}}|$ and denote the node-feature matrix at time step $t$ as
$\mathbf{X}_t\ = \big[\mathbf{x}_u\big]_{u=1}^{N} \in \mathbb{R}^{N\times d_{in}}$.
We first apply a one-layer MLP to map the input features to the model hidden width $d$:
\begin{equation}
\mathbf{H}_t^{(0)}=\mathrm{MLP}_{\text{in}}(\mathbf{X}_t)=\sigma(\mathbf{X}_t \mathbf{W}_{in}+\mathbf{b}_{in})\in\mathbb{R}^{N\times d},
\end{equation}
where $\mathbf{W}_{in}\in\mathbb{R}^{d_{in}\times d}$ and $\sigma(\cdot)$ is a point-wise nonlinearity.

%% step 2. 线性化的核心
%% 创新点：结合邻接矩阵 A，构造了 structure-aware 的一种先验投影，融合到了 SGFormer 里。
\subsubsection{Structure-aware global mixing via linear attention.}
Standard softmax self-attention incurs quadratic time in the number of node\-s, which quickly becomes a bottleneck on large span graphs.
Since our latency predictor runs online in an auto-scaling control loop, inference latency must remain low as $N$ grows.
Prior work reduces the attention cost by replacing dot-product attention with alternative mixing operators~\cite{zhai2021aft} or by restricting the attended set via sparse patterns~\cite{longformer}.
In our setting, each node should still aggregate information from all other nodes within the graph snapshot to capture global dependency propagation.
Inspired by SGFormer~\cite{sgformer} and early linear-attention designs for Transformers~\cite{katharopoulos2020lineartransformer,hua2022lineartransformer}, we implement a linear global mixing operator that enables dense all-pair mixing without constructing the $N\times N$ attention matrix.
Meanwhile, each API $\mathcal{A}$ induces a time-invariant sparse topology, which provides a strong structural prior for dependency propagation on trace-induced span graphs.
We inject this topology into global mixing via a one-shot sparse pre-mixing step on keys and values.
We compute \emph{query}/\emph{key}/\emph{value} projections:
\begin{equation}
\mathbf{Q}=\sigma(\mathbf{H}_t^{(0)}\mathbf{W}_Q),\quad
\mathbf{K}=\sigma(\mathbf{H}_t^{(0)}\mathbf{W}_K),\quad
\mathbf{V}=\mathbf{H}_t^{(0)}\mathbf{W}_V,
\end{equation}
where $\mathbf{W}_Q,\mathbf{W}_K,\mathbf{W}_V\in\mathbb{R}^{d\times d}$ and $\sigma(\cdot)$ is a non-negative activation.
We use $\mathrm{ReLU}(\cdot)$ so that the diagonal normalization $\mathbf{D}_t$ remains positive and well-conditioned.
To make the global mixing insensitive to feature scale and varying graph sizes across APIs and windows, we follow SGFormer and normalize queries and keys by the Frobenius norm:
\begin{equation}
\tilde{\mathbf{Q}}=\frac{\mathbf{Q}}{\|\mathbf{Q}\|_{\mathcal{F}} + \epsilon},\quad
\tilde{\mathbf{K}}=\frac{\mathbf{K}}{\|\mathbf{K}\|_{\mathcal{F}} + \epsilon},
\end{equation}
where $\|\cdot\|_{\mathcal{F}}$ denotes the Frobenius norm and $\epsilon$ is a small constant for numerical stability.

\paragraph{Sparse structure-aware pre-mixing (one-shot).}
Let $\mathbf{A}\in\{0,1\}^{N\times N}$ be the adjacency matrix of $\mathcal{E}^{\mathcal{A}}$ constructed in Algorithm~\ref{alg:span-graph-construction}.
We add self-loops and construct a row-stochastic propagation operator:
\begin{equation}
\hat{\mathbf{A}}=\mathbf{A}+\mathbf{I},\quad
\hat{\mathbf{D}}=\mathrm{diag}(\hat{\mathbf{A}}\mathbf{1}),\quad
\mathbf{P}=\hat{\mathbf{D}}^{-1}\hat{\mathbf{A}}.
\end{equation}
We then apply one-shot pre-mixing controlled by a scalar $\rho\in[0,1]$:
\begin{equation}
\bar{\mathbf{K}}=(1-\rho)\tilde{\mathbf{K}}+\rho\,\mathbf{P}\tilde{\mathbf{K}},\quad
\bar{\mathbf{V}}=(1-\rho)\mathbf{V}+\rho\,\mathbf{P}\mathbf{V}.
\label{eq:structure-aware-premix}
\end{equation}
Since $\mathbf{P}$ is time-invariant for a fixed API, it can be precomputed and applied once per window.
We apply Eq.~\eqref{eq:structure-aware-premix} only once to avoid iterative message passing and keep inference lightweight.

\paragraph{Linear global mixing with structure-informed summaries.}
Let $\mathbf{1}\in\mathbb{R}^{N}$ be the all-ones vector, and define the diagonal normalization matrix
\begin{equation}
\mathbf{D}_{t}=\mathrm{diag}\!\left(\mathbf{1}+\frac{1}{N}\tilde{\mathbf{Q}}(\bar{\mathbf{K}}^\top \mathbf{1})\right)\in\mathbb{R}^{N\times N}.
\end{equation}

Then the global representations are computed by
\begin{equation}
\mathbf{Z}^{G}_t
=
\mathbf{D}_t^{-1}\!\left[\bar{\mathbf{V}}+\frac{1}{N}\tilde{\mathbf{Q}}(\bar{\mathbf{K}}^\top \bar{\mathbf{V}})\right]\in\mathbb{R}^{N\times d}.
\label{eq:global-linear-attn}
\end{equation}

Eq.~\eqref{eq:global-linear-attn} avoids constructing the $N\times N$ attention matrix by computing $\bar{\mathbf{K}}^\top \bar{\mathbf{V}}\in\mathbb{R}^{d\times d}$ and $\bar{\mathbf{K}}^\top \mathbf{1}\in\mathbb{R}^{d}$ once, yielding $\mathcal{O}(N d^2)$ time for fixed width $d$.
Together with the one-shot sparse pre-mixing (Eq.~\eqref{eq:structure-aware-premix}), the overall global mixing costs $\mathcal{O}(|\mathcal{E}^{\mathcal{A}}|d+N d^2)$ and introduces no $\mathcal{O}(N^{2})$ memory term.
Intuitively, $\bar{\mathbf{K}}^\top \bar{\mathbf{V}}$ is a structure-informed global summary of values across all nodes, and left-multi\-plying it by $\tilde{\mathbf{Q}}$ distributes this summary to each node; the diagonal matrix $\mathbf{D}_t$ provides per-node normalization.
This realizes all-pair interaction in linear time, which is important in our setting: tail latency is often dominated by rare long traces with deep and branching call paths, and these cases are also where quadratic attention can become the bottleneck.

\subsubsection{Local propagation via a GCN (GN branch).}
To capture local dependency propagation on $\mathcal{G}^{\mathcal{A}}$, we apply a single graph convolution (GCN) layer.
Using the self-looped adjacency $\hat{\mathbf{A}}$ and degree matrix $\hat{\mathbf{D}}$ from the pre-mixing step, the local representations are:
\begin{equation}
\mathbf{Z}^{L}_t = \hat{\mathbf{D}}^{-\frac12}\hat{\mathbf{A}} \hat{\mathbf{D}}^{-\frac12} \mathbf{H}_t^{(0)} \mathbf{W}_L \in\mathbb{R}^{N\times d},
\label{eq:local-diffusion}
\end{equation}
where $\mathbf{W}_L\in\mathbb{R}^{d\times d}$.
Since $\hat{\mathbf{A}}$ is sparse and time-invariant for a fixed call path, Eq.~\eqref{eq:local-diffusion}
runs in $\mathcal{O}(|\mathcal{E}^{\mathcal{A}}|d)$ and can reuse the precomputed normalization.

\paragraph{Fusion.}
We fuse global and local representations with a row-wise feed-forward network and a residual connection.
Let $\mathbf{Z}_t=[\mathbf{Z}^{G}_t \parallel \mathbf{Z}^{L}_t]\in\mathbb{R}^{N\times 2d}$.
We compute a fused update by a two-layer MLP:
\begin{equation}
\Delta\mathbf{H}_t
=
\sigma(\mathbf{Z}_t\mathbf{W}_{f,1}+\mathbf{b}_{f,1})\mathbf{W}_{f,2}+\mathbf{b}_{f,2}\in\mathbb{R}^{N\times d},
\end{equation}
and obtain the final node representations by
\begin{equation}
\mathbf{H}_t=\mathrm{LN}\!\left(\mathbf{H}_t^{(0)}+\Delta\mathbf{H}_t\right)\in\mathbb{R}^{N\times d},
\label{eq:fusion}
\end{equation}
where $[\cdot\parallel\cdot]$ denotes concatenation, $\mathrm{LN}(\cdot)$ is layer normalization applied to each node representation,
$\mathbf{W}_{f,1}\in\mathbb{R}^{2d\times d}$ and $\mathbf{W}_{f,2}\in\mathbb{R}^{d\times d}$ are shared across nodes.
The fusion is row-wise and does not introduce any $\mathcal{O}(N^2)$ term.

\paragraph{Readout and context fusion.}
We first aggregate node representations into a graph embedding $\mathbf{g}^{\mathrm{node}}_t\in\mathbb{R}^{d}$ via attention pooling:
\begin{equation}
\alpha_{t,i}=\frac{\exp(\mathbf{w}_r^\top \mathbf{h}_{t,i})}{\sum_{j=1}^N \exp(\mathbf{w}_r^\top \mathbf{h}_{t,j})},\quad
\mathbf{g}^{\mathrm{node}}_t=\sum_{i=1}^N \alpha_{t,i} \mathbf{h}_{t,i},
\end{equation}
where $\mathbf{h}_{t,i}$ is the $i$-th row of $\mathbf{H}_t$ and $\mathbf{w}_r\in\mathbb{R}^{d}$ is a learnable vector shared across windows.
We then incorporate the trace-common context vector $\mathbf{c}_t\in\mathbb{R}^{d_c}$ constructed from $\mathcal{T}^{\mathcal{A}}_{t}$ in Sec.~\ref{sec:data-collector} to form the window representation used by the temporal decoder:
\begin{equation}
\bar{\mathbf{c}}_t=\sigma(\mathbf{c}_t\mathbf{W}_c+\mathbf{b}_c)\in\mathbb{R}^{d},\ 
\mathbf{g}_t=\sigma\!\left([\mathbf{g}^{\mathrm{node}}_t\parallel \bar{\mathbf{c}}_t]\mathbf{W}_g+\mathbf{b}_g\right)\in\mathbb{R}^{d},
\label{eq:readout}
\end{equation}
where $\mathbf{W}_c\in\mathbb{R}^{d_c\times d}$ and $\mathbf{W}_g\in\mathbb{R}^{2d\times d}$.
This design summarizes node-level information first, then fuses trace-level context to provide a compact feature for temporal modeling.

\subsection{Temporal Decoder}

Given the history length $L$, we stack the last-$L$ window representations into a multivariate time series
$\mathbf{G}_t=[\mathbf{g}_{t-L+1};\ldots;\mathbf{g}_t]\in\mathbb{R}^{L\times d}$.
The tail-latency series contains both periodic workload cycles and aperiodic bursts.
To model such mixed temporal patterns, we adopt a temporal decoder based on the TimesBlock module of TimesNet~\cite{timesnet},
which captures multi-frequency temporal variations by extracting dominant periods from the frequency domain and aggregating period-wise representations.
We first extend the input to length $T=L+H$ by appending $H$ placeholder slots (zero padding), then apply a one-layer MLP to each time step:
\begin{equation}
\mathbf{U}_t^{(0)} = \mathrm{MLP}_{\text{ts}}\bigl([\mathbf{G}_t;\mathbf{0}_{H\times d}]\bigr)\in\mathbb{R}^{T\times d},
\end{equation}
where $\mathrm{MLP}_{\text{ts}}(\cdot)$ is applied row-wise and uses the identity activation.
We then apply $B$ stacked TimesBlocks with residual connections:
\begin{equation}
\mathbf{U}_t^{(b+1)} = \mathbf{U}_t^{(b)} + \mathrm{TimesBlock}\!\left(\mathbf{U}_t^{(b)}\right),\ b=0,\ldots,B-1.
\label{eq:timesnet-stack}
\end{equation}

\paragraph{TimesBlock.}
For a given input $\mathbf{U}\in\mathbb{R}^{T\times d}$, TimesBlock identifies $K$ dominant periods
$\mathcal{P}=\{p_1,\ldots,p_K\}$ from the FFT spectrum and constructs $K$ period-wise representations $\{\mathbf{U}_i\}_{i=1}^{K}$.
Each $\mathbf{U}_i$ is obtained by padding and reshaping $\mathbf{U}$ into a 2D representation, applying a 2D convolutional block, and reshaping it back, following TimesNet.
This provides a multi-period inductive bias for capturing periodic components at different frequencies, while the residual stacking in Eq.~\eqref{eq:timesnet-stack} preserves flexibility for aperiodic deviations.
The period-wise representations are then aggregated with softmax-normalized weights $\omega_i$ derived from FFT amplitudes:
\begin{equation}
\mathrm{TimesBlock}(\mathbf{U})=\sum_{i=1}^{K}\omega_i \mathbf{U}_i,\ 
\omega_i=\frac{\exp(a_i)}{\sum_{j=1}^K \exp(a_j)}.
\label{eq:timesblock-agg}
\end{equation}

\paragraph{Prediction head.}
The decoder output $\mathbf{U}_t^{(B)}\in\mathbb{R}^{T\times d}$ contains both reconstructed history and prediction slots.
We map the last $H$ slots to the target tail latency predictions:
\begin{equation}
\hat{\mathbf{y}}_{t+1:t+H} = \mathrm{MLP}_{\text{out}}\!\left(\mathbf{U}_t^{(B)}[L+1:L+H]\right)\in\mathbb{R}^{H},
\label{eq:prediction-head}
\end{equation}
where $\mathrm{MLP}_{\text{out}}(\cdot)$ is applied row-wise to map each prediction slot to a scalar latency.

\subsection{Quantile Loss}

We focus on p95 latency prediction in this paper and set $q=0.95$.
We train STLGT using the quantile regression (pinball) loss.
Let $y_{t+h}$ be the ground-truth tail latency at horizon $h$ and $\hat{y}_{t+h}$ be the prediction.
The per-sample quantile loss is defined as
\begin{equation}
\ell_q(y,\hat y)=\psi_q(y-\hat y),\ 
\psi_q(u)=\max(qu,(q-1)u).
\end{equation}
For high quantiles, this asymmetric loss penalizes underestimation more than overestimation, aligning with the goal of tail-latency prediction.
Other latency percentiles (such as p50 and p90) are used as input features in the trace-common context $\mathbf{c}_t$.
Over a training set of time indices $\mathcal{T}$ and horizons $h=1,\ldots,H$, the objective is
\begin{equation}
\mathcal{L}=\frac{1}{|\mathcal{T}|H}\sum_{t\in\mathcal{T}}\sum_{h=1}^{H}\ell_q\!\left(y_{t+h},\ \hat{y}_{t+h}\right).
\label{eq:quantile-loss}
\end{equation}

%% file: sections/5-experiments.tex
\section{Experiments}
\label{sec:experiments}

%% RQ definition
To evaluate STLGT (Fig.~\ref{fig:stlgt}) and its key design components (Fig.~\ref{fig:encoder}), we conduct experiments to answer the following research questions:

\noindent\textbf{RQ1: Accuracy.}
How accurate is STLGT for multi-step p95 tail-latency prediction compared with representative baselines under our mixed workload setting and across datasets?

\noindent\textbf{RQ2: Ablation.}
How do the main components of STLGT (Structur\-e-aware global mixing, local propagation, trace-aware readout, and temporal decoder) contribute to prediction accuracy?

\noindent\textbf{RQ3: Efficiency.}
How efficient is STLGT for online inference on production-scale span graphs, and how does inference latency scal\-e with the number of stage nodes $N$ and hidden width $d$?

\noindent\textbf{RQ4: Education Applicability.}
Can STLGT provide accurate and stable prediction for personalized education microservices, especially during exam-oriented traffic bursts?

\subsection{Experiment Settings}
\label{sec:exp-settings}

\subsubsection{Benchmarks and Datasets}
\label{sec:exp-datasets}

\paragraph{DeathStarBench microservices.}
We evaluate STLGT on two open-source microservice applications from DeathStarBench (DSB)~\cite{deathstarbench}:
\emph{Hotel Reservation} (17 microservices) and \emph{Social Network} (26 micro\-services).
We enable distributed tracing via Istio (sidecar-based context propagation) with Jaeger as the tracing backend,
and collect per-service metrics using Prometheus and cAdvisor together with per-request traces.
Following Sec.~\ref{sec:data-collector}, we aggregate observations into fixed time windows of length $\Delta=30$s.
For each application, we select five API endpoints and use their dominant call paths for trace collection and model training.
We train and evaluate models on each trace stream independently, and report the mean results across the five streams
to mitigate endpoint-specific bias.

\paragraph{Alibaba production traces.}
We additionally evaluate on the Alibaba trace dataset~\cite{ali-traces}, which provides anonymized production traces with service dependencies and end-to-end latencies.
Du et al.~\cite{du2025generating} re-analyze this corpus and report that the per-trace call subset size ranges from 1 to 10.
Accordingly, we select six subset sizes $\{1,2,4,6,8,10\}$ to cover this range and sample one trace subset for each size.
Since the service-level metrics in the Alibaba traces are recorded at 60-second intervals, we set $\Delta=60$s.
For each subset, we use a 24-hour trace slice, yielding 1{,}440 windows.
We follow the same feature construction procedure as in Sec.~\ref{sec:data-collector}.

\paragraph{Edu Platform.}
We further evaluate STLGT on an internal personalized education service platform.
The platform supports four basic-education subjects and five learning scenarios, covering heterogeneous teaching activities and learner interactions.
Consistent with the DeathStarBench setting, we select two representative teaching scenarios for targeted evaluation, including an exam-oriented scenario where many students access authentication, question-bank, submission, and grading services within a narrow time interval.
This setting is representative of education platforms because request bursts are often scheduled by course and exam timetables rather than emerging randomly from external user behavior.

\begin{table}[t]
  \centering
  \caption{Dataset statistics. \#Win counts window-level samples aggregated over selected trace streams (DSB) or sampled trace subsets (Alibaba).
  For Alibaba traces, \#SVC counts the total number of microservice nodes in the cluster (20K+), while the call subset size per trace ranges from 1 to 10 in~\cite{ali-traces}; we sample six representative subset sizes from this range.}
  \label{tab:dataset-stats}
  \vspace{2pt}
  \begin{tabular}{l|c|c|c|c}
    \toprule
    \textbf{Dataset} & \textbf{\#SVC} & \textbf{\#Streams} & $\Delta$ & \textbf{\#Win} \\
    \midrule
    Hotel Reservation (DSB) & 17 & 5 & 30s & 14{,}400 \\
    Social Network (DSB) & 26 & 5 & 30s & 14{,}400 \\
    Alibaba traces & 20K+ & 6 & 60s & 8{,}640 \\
        \bottomrule
  \end{tabular}
\end{table}

\paragraph{Prediction target.}
For each call path $\mathcal{A}$ and window $t$, the label is the p95 end-to-end trace latency in that window:
$y_t^{\mathcal{A}}=\rho_{0.95}(\mathcal{T}_t^{\mathcal{A}})$.
Unless otherwise stated, we set history length $L=12$ and horizon $H=6$, i.e., predict the next $H$ windows from the past $L$ windows.

\subsubsection{Workload Pattern (DeathStarBench)}
\label{sec:exp-workloads}

\begin{figure}[ht]
  \centering
  \includegraphics[width=\linewidth]{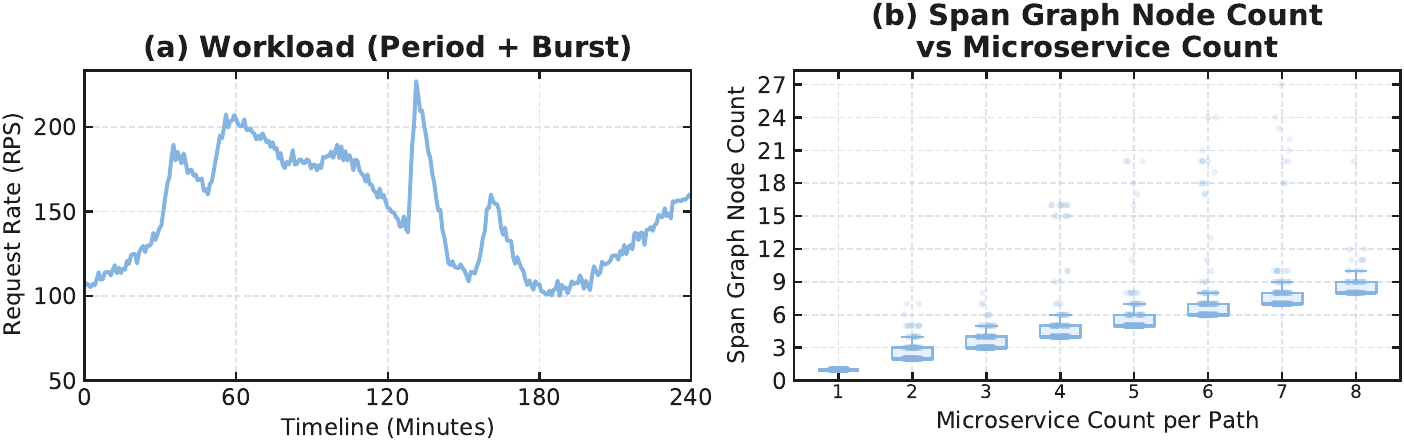}
  \Description{(a) A request-rate time series showing a periodic baseline with injected short bursts. (b) A scatter/boxplot summarizing the distribution of span graph node count versus microservice count per request/path in Alibaba production traces.}
  \caption{Workload and production-trace characteristics. (a) Example request-rate trace generated by Locust with a periodic (cosine) baseline plus short burst injections. (b) Empirical distribution of span graph node count versus microservice count per request/path in Alibaba production traces.}
  \label{fig:workload}
\end{figure}

We drive the microservices with a single \emph{mixed} workload generated by Locust~\cite{locust}.
This workload combines a steady baseline, periodic variation, and short burst injections to emulate realistic production request-rate dynamics.
We schedule a target request rate $r(t)$ over time as
$r(t)=r_{0}+a\cos(2\pi t/P)+\Delta r(t)$,
where $r_{0}$ sets the nominal load level and the cosine term models regular fluctuations such as diurnal effects.
On top of this periodic baseline, we inject burst segments $\Delta r(t)$ at random start times to mimic non-periodic yet often event-driven surges (e.g., scheduled flash-sale campaigns).
In Locust, each burst is implemented by temporarily spawning additional distributed us\-ers for a short interval; we intentionally keep bursts brief and truncate their tails to create abrupt rises and drops in the realized request rate.
Fig.~\ref{fig:workload}(a) illustrates an example trace.

For each application, we run 24 hours of continuous testing, yielding 2{,}880 windows (with $\Delta=30$s).
We discard warm-up windows and windows with missing/invalid traces.
The window coun\-ts in Table~\ref{tab:dataset-stats} are aggregated over selected call paths.
To bound tracing overhead, we uniformly sub-sample up to 150 traces per call path in each window for feature aggregation, yielding 432{,}000 sampled traces per call path per day.

\subsubsection{Implementation and Hardware}
\label{sec:exp-setup}

We deploy a Kubernetes clu\-ster on three cloud instances running Ubuntu 22.04, each with 12 CPU cores and 24GB of memory, and run the microservice applications on this cluster.
We use a single NVIDIA TITAN RTX GPU for model training and GPU inference benchmarking.
Unless otherwise stated, STLGT uses hidden width $d=32$ (to match baselines), one spatial encoder block as described in Sec.~\ref{sec:latency-predictor}, and a TimesNet decoder with $B=2$ stacked TimesBlocks and $K=3$ dominant periods.

\subsubsection{Baselines and Variants}
\label{sec:exp-baselines}

\paragraph{Baselines.}
We compare STLGT with a non-graph regression ba\-seline GBDT~\cite{gbdt} and representative trace-graph baselines:
PERT-GNN~\cite{pert} and FastPERT~\cite{fast-pert}.
All baselines are trained on the same raw data; when a baseline requires a different graph abstraction (e.g., PERT-style multi-stage splitting),
we follow the original construction and only adapt the prediction head to output $H$-step-ahead p95 latency.

\paragraph{Ablations.}
We evaluate STLGT variants by removing or replacing key components:
(0) \textbf{Standard Attention}: replace the structu\-re-aware linear attention with standard softmax self-attention;
(1) \textbf{w/o Structure-aware}: remove the one-shot sparse pre-mixing in Eq.~\eqref{eq:structure-aware-premix} (set $\rho=0$);
(2) \textbf{w/o Local}: remove the GCN branch in Eq.~\eqref{eq:local-diffusion};
(3) \textbf{w/o Trace-context}: remove $\mathbf{c}_t$ from the readout in Eq.~\eqref{eq:readout};
(4) \textbf{Linear decoder}: replace TimesNet with a linear temporal head.

\subsubsection{Training Details and Evaluation Metrics}
\label{sec:exp-protocol}

We use chronological splits (70\%/15\%/15\%) to avoid temporal leakage.
We repeat each experiment five times with different random seeds and report the mean and standard deviation.
Except for GBDT, all neural models are trained with AdamW (learning rate $10^{-3}$, weight decay $10^{-4}$), batch size 32, and early stopping on the validation MAE.
GBDT is trained with its standard gradient-boosted-tree objective under the same chronological split, and the validation set is used for model selection.

\paragraph{Accuracy metrics.}
We report mean absolute error (MAE) and mean absolute percentage error (MAPE) averaged over the prediction horizon $h=1,\dots,H$.

\paragraph{Efficiency metrics.}
We report inference latency (ms per window) and throughput (windows/s) for forward passes with batch size 1,
measured as the average over 1,000 runs after warm-up.

\subsection{Experiment Results}
\label{sec:exp-results}

\subsubsection{Overall Accuracy Across Datasets (RQ1)}
\label{sec:exp-rq1}

We compare STLGT against GBDT, PERT-GNN, and FastPERT on two DeathStarBench applications, Alibaba production traces, and the Edu Platform. Table~\ref{tab:main-results} reports the overall results (rows 1--4) along with ablations, where each cell shows the mean and standard deviation over multiple runs.

Overall, STLGT consistently outperforms all baselines across all four datasets. It achieves MAPE of 10.12\% on Social Network, 9.89\% on Hotel Reservation, 9.60\% on Alibaba traces, and 9.99\% on Edu Platform. Compared with PERT-GNN, STLGT reduces MAPE by 9.5\%, 9.1\%, 7.0\%, and 8.3\%, respectively, corresponding to an average improvement of 8.5\%.

Furthermore, the reported standard deviations indicate that these gains are stable and not attributable to random initialization or training noise. Beyond relative error, STLGT also achieves the best MAE among full models on Social Network, Hotel Reservation, and Alibaba traces, further confirming its advantage.

These improvements align with the design of STLGT. The spatial encoder captures long-range service dependencies through structure-aware global mixing and local propagation, while the temporal decoder models workload evolution across time windows. Under mixed workloads that combine periodic patterns with bursty spikes, this spatiotemporal design enables STLGT to better capture both cross-service latency propagation and rapid fluctuations in request intensity.

We also observe a different pattern on the Edu Platform: STLGT achieves the best MAPE, while FastPERT attains a lower MAE. This indicates that STLGT provides stronger relative-error control across latency levels, whereas FastPERT fits the dominant low-latency range more aggressively. For education workloads, this distinction is important because routine learning traffic and exam bursts may occupy different latency regimes; the lower MAPE suggests that STLGT is less biased toward only the most frequent operating range.

\newcommand{\second}[1]{\smash{\underline{#1}}\vphantom{#1}}
\newcommand{\resultcell}[2]{\shortstack[c]{\rule{0pt}{1.05em}#1\\[-1pt]{\scriptsize\strut($\pm$#2)}}}

\begin{table*}[t]
  \centering
  \caption{Overall accuracy and ablation results. Each cell reports mean $\pm$ standard deviation over repeated runs. Best is in bold and second-best is underlined.}
  \label{tab:main-results}
  \vspace{2pt}
  \small
  \renewcommand{\arraystretch}{1.08}
  \setlength{\tabcolsep}{3.6pt}
  \begin{tabular}{@{}lcccccccc@{}}
    \toprule
    \multirow{2}{*}{\textbf{Method}} &
    \multicolumn{2}{c}{\shortstack{\textbf{Social Network}}} &
    \multicolumn{2}{c}{\shortstack{\textbf{Hotel Reservation}}} &
    \multicolumn{2}{c}{\shortstack{\textbf{Alibaba Traces}}} &
    \multicolumn{2}{c}{\shortstack{\textbf{Edu Platform}}} \\
    \cmidrule(lr){2-3}\cmidrule(lr){4-5}\cmidrule(lr){6-7}\cmidrule(lr){8-9}
    & \textbf{MAPE (\%)} & \textbf{MAE (ms)}
    & \textbf{MAPE (\%)} & \textbf{MAE (ms)}
    & \textbf{MAPE (\%)} & \textbf{MAE (ms)}
    & \textbf{MAPE (\%)} & \textbf{MAE (ms)} \\
    \midrule
    (1) GBDT~\cite{gbdt} 
      & \resultcell{21.73\%}{0.38\%}
      & \resultcell{40.13}{0.27}
      & \resultcell{19.10\%}{0.34\%}
      & \resultcell{34.02}{0.31}
      & \resultcell{19.37\%}{0.39\%}
      & \resultcell{13.43}{0.32}
      & \resultcell{20.05\%}{0.33\%}
      & \resultcell{1.60}{0.08} \\
    (2) PERT-GNN~\cite{pert}
      & \resultcell{11.18\%}{0.44\%}
      & \resultcell{21.37}{0.38}
      & \resultcell{10.88\%}{0.31\%}
      & \resultcell{19.48}{0.26}
      & \resultcell{10.32\%}{0.39\%}
      & \resultcell{7.03}{0.31}
      & \resultcell{10.89\%}{0.27\%}
      & \resultcell{1.24}{0.16} \\
    (3) FastPERT~\cite{fast-pert}
      & \resultcell{11.59\%}{0.27\%}
      & \resultcell{21.51}{0.20}
      & \resultcell{10.82\%}{0.28\%}
      & \resultcell{19.14}{0.15}
      & \resultcell{10.86\%}{0.22\%}
      & \resultcell{7.57}{0.13}
      & \resultcell{10.96\%}{0.26\%}
      & \resultcell{\second{0.71}}{0.05} \\
    (4) \textbf{STLGT (Ours)}
      & \resultcell{\textbf{10.12\%}}{0.16\%}
      & \resultcell{\textbf{18.93}}{0.14}
      & \resultcell{\second{9.89\%}}{0.15\%}
      & \resultcell{\second{17.72}}{0.13}
      & \resultcell{\textbf{9.60\%}}{0.14\%}
      & \resultcell{\textbf{6.25}}{0.09}
      & \resultcell{\textbf{9.99\%}}{0.12\%}
      & \resultcell{1.42}{0.07} \\
    \midrule
    \multicolumn{9}{@{}c@{}}{\textbf{Ablation Study}} \\
    \midrule
    (5) STLGT w/ Standard Attention
      & \resultcell{\second{10.30\%}}{0.18\%}
      & \resultcell{\second{18.96}}{0.15}
      & \resultcell{\textbf{9.86\%}}{0.14\%}
      & \resultcell{\textbf{17.07}}{0.12}
      & \resultcell{10.24\%}{0.17\%}
      & \resultcell{7.14}{0.10}
      & \resultcell{10.21\%}{0.15\%}
      & \resultcell{0.99}{0.06} \\
    (6) STLGT w/o Structure-aware ($\rho=0$)
      & \resultcell{10.73\%}{0.21\%}
      & \resultcell{19.78}{0.17}
      & \resultcell{10.69\%}{0.19\%}
      & \resultcell{18.41}{0.16}
      & \resultcell{10.70\%}{0.18\%}
      & \resultcell{7.01}{0.11}
      & \resultcell{11.03\%}{0.17\%}
      & \resultcell{\textbf{0.68}}{0.04} \\
    (7) STLGT w/o Local (GCN)
      & \resultcell{12.14\%}{0.28\%}
      & \resultcell{23.03}{0.23}
      & \resultcell{11.88\%}{0.24\%}
      & \resultcell{20.93}{0.20}
      & \resultcell{11.81\%}{0.21\%}
      & \resultcell{7.78}{0.14}
      & \resultcell{12.16\%}{0.20\%}
      & \resultcell{0.94}{0.05} \\
    (8) STLGT w/o Trace-context ($\mathbf{c}_t$)
      & \resultcell{10.57\%}{0.19\%}
      & \resultcell{19.82}{0.16}
      & \resultcell{10.39\%}{0.17\%}
      & \resultcell{18.10}{0.14}
      & \resultcell{10.04\%}{0.16\%}
      & \resultcell{7.10}{0.10}
      & \resultcell{10.51\%}{0.14\%}
      & \resultcell{1.01}{0.06} \\
    \shortstack[l]{(9) STLGT w/ Linear Decoder}
      & \resultcell{12.41\%}{0.31\%}
      & \resultcell{23.68}{0.25}
      & \resultcell{11.91\%}{0.26\%}
      & \resultcell{21.02}{0.21}
      & \resultcell{12.38\%}{0.24\%}
      & \resultcell{8.64}{0.15}
      & \resultcell{12.43\%}{0.22\%}
      & \resultcell{1.22}{0.07} \\
    \bottomrule
  \end{tabular}
\end{table*}

\subsubsection{Ablation Study and Design Choices (RQ2)}
\label{sec:exp-rq2}

Table~\ref{tab:main-results} reports ablation results (rows 5--9) by removing or replacing key components of STLGT.

We first examine the effect of the \emph{structure-aware} linear attention. Compared with standard softmax attention (row 5), it achieves better performance on Social Network and Alibaba traces, while remaining within 1\% on Hotel Reservation, indicating that linearizing attention with topology-aware priors preserves accuracy in practice.

Removing either the structure-aware pre-mixing ($\rho=0$, row 6) or the trace-context fusion $\mathbf{c}_t$ (row 8) consistently degrades performance, highlighting the importance of both structural bias and trace-level contextual information.

Among all variants, removing local propagation (GCN, row 7) leads to the largest performance drop, suggesting that capturing local dependency patterns is critical. In addition, replacing TimesNet with a linear temporal head (row 9) further hurts performance, confirming the necessity of effective temporal modeling under mixed workloads with burst injections.

Overall, these results demonstrate that both spatial dependency modeling and temporal dynamics are indispensable for accurate tail-latency prediction.

\subsubsection{Hidden Width Selection Across DSB and Edu (RQ2 and RQ4)}
\label{sec:exp-d}

\begin{figure}[ht]
  \centering
  \includegraphics[width=\linewidth]{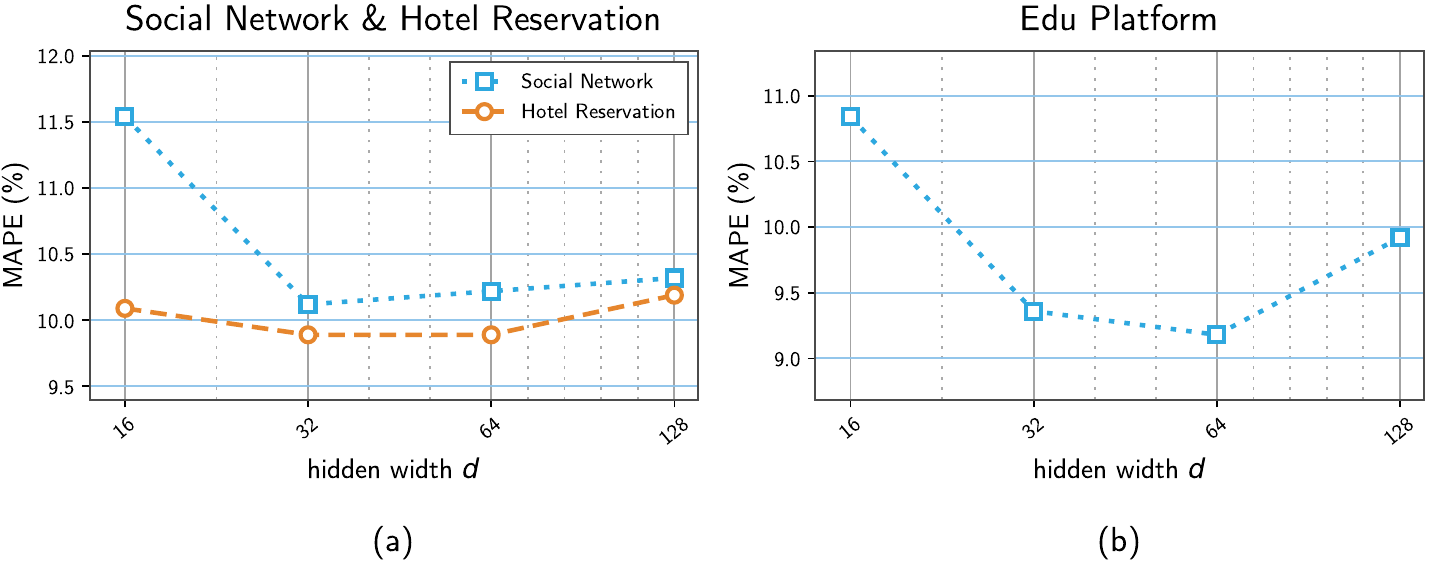}
  \Description{Prediction MAPE versus hidden width d. Panel (a) reports the DSB benchmarks, including Social Network and Hotel Reservation, and panel (b) reports the Edu Platform. Lower is better.}
  \caption{Effect of hidden width $d$ on mean MAPE: (a) DSB microservice benchmarks, including Social Network and Hotel Reservation; (b) Edu Platform. Lower is better.}
  \label{fig:hidden-width}
\end{figure}

We vary the hidden width $d \in \{16, 32, 64, 128\}$ while keeping all other settings fixed.
Fig.~\ref{fig:hidden-width} reports the mean MAPE averaged over multiple runs on (a) the DSB microservice benchmarks, including Social Network and Hotel Reservation, and (b) the Edu Platform.

Across both benchmark and education settings, $d=16$ is too narrow and leads to noticeably higher error, indicating that the model needs sufficient capacity to encode dependency propagation and temporal dynamics.
Moving to a moderate width ($d=32$ or $64$) captures most of the gain: the DSB curves become stable after the hidden representation is sufficiently expressive, while the Edu Platform shows the same preference for a moderate hidden width under education-specific workloads.
Increasing the width further to $d=128$ does not provide consistent benefits and may add optimization difficulty and inference cost without introducing useful signal.
We therefore use $d=32$ as the default because it achieves near-best accuracy while keeping online inference lightweight.

\subsubsection{Inference Scalability (RQ3)}
\label{sec:exp-rq3}

We first analyze the Alibaba production traces to understand the span graph regime in practice.
As shown in Fig.~\ref{fig:workload}(b), each trace involves no more than 10 microservices and induces span graphs with up to $N=32$ stage nodes, with a clear long-tail distribution (most graphs are small but some approach the upper end).
Moreover, most span graphs are tree-like, so the edge set is sparse (i.e., $|\mathcal{E}|=\mathcal{O}(N)$).

The goal of STLGT is therefore not to be significantly faster than PERT-GNN on tiny span graphs, but to improve \emph{scalability} and reduce the gap between short-chain and long-chain traces as $N$ grows.
This scaling benefit mainly comes from the linear-attention-based spatial encoder, which avoids materializing an $N\times N$ attention matrix.
This matters in production: a scheduler typically queries the predictor hundreds or thousands of times per control interval (e.g., across APIs, candidate actions, or trace subsets), so a few large-$N$ cases can dominate the end-to-end decision latency.
Table~\ref{tab:efficiency} compares complexity, model size, and batch-1 inference latency under two representative graph sizes ($N\!=\!1$ and $N\!=\!32$) with $d=32$.
Both models have similar latency at $N=1$, but PERT-GNN scales much more steeply, especially on CPU, due to its $\mathcal{O}(N^2 d)$ mixing.
Although STLGT includes multiple projection matrices and a TimesBlock-based temporal decoder, most parameters are in dense linear/conv layers with high parallelism; in practice, the scalability gap is dominated by the spatial mixing cost.
From $N=1$ to $N=32$, STLGT shows much smaller latency growth (1.5$\times$ on GPU and 3.8$\times$ on CPU) than PERT-GNN (2.0$\times$ and 49.7$\times$, respectively).

\begin{table*}[t]
  \centering
  \caption{Inference scalability comparison between STLGT and PERT-GNN with $d=32$. We report asymptotic complexity, model size (\#Params), and batch-1 inference latency under two span graph sizes ($N=1$ and $N=32$, corresponding to the small-graph and upper-tail regimes in Fig.~\ref{fig:workload}(b)). We also report the latency growth from $N=1$ to $N=32$ (ratio). Lower growth indicates better scalability and smaller tail impact when the predictor is invoked many times per scheduling decision.}
  \label{tab:efficiency}
  \vspace{2pt}
  \small
  \begin{tabular}{l|c|c|c|c}
    \toprule
    \textbf{Model} & \textbf{Complexity} & \textbf{\#Params} & \textbf{GPU inference latency (ms)} & \textbf{CPU inference latency (s)} \\
    \midrule
    PERT-GNN~\cite{pert} & $\mathcal{O}(N^2 d)$ & 0.12M & 30.3 $\rightarrow$ 60.72 (2.0$\times$) & 0.21 $\rightarrow$ 10.44 (49.7$\times$) \\
    \textbf{STLGT (Ours)} & $\mathcal{O}(|\mathcal{E}|d + N d^2)$ & 0.05M & 30.5 $\rightarrow$ 46.0 (1.5$\times$) & 0.22 $\rightarrow$ 0.84 (3.8$\times$) \\
    \bottomrule
  \end{tabular}
\end{table*}

\subsubsection{Case Study: Edu Platform Under Exam Bursts (RQ4)}
\label{sec:exp-edu-case}

The Edu Platform represents a practical deployment context where workload bursts are not merely random spikes but are often induced by teaching schedules.
During online exams, many students enter the system almost simultaneously, repeatedly access question-bank and submission APIs, and trigger downstream grading or analytics services within a short interval.
This produces a sharp increase in request intensity and can amplify tail latency through shared services such as authentication, course management, question retrieval, submission persistence, and score processing.
Accurate p95 prediction in this setting is therefore useful for proactive scaling because a delayed response may directly affect exam fairness and user experience.

Table~\ref{tab:main-results} shows that STLGT achieves the lowest MAPE on the Edu Platform (9.99\%), improving over PERT-GNN by 8.3\% and outperforming both GBDT and FastPERT in relative-error accuracy.
This result indicates that STLGT can track the proportional change of tail latency across routine learning periods and short exam-induced bursts.
The advantage mainly comes from two design choices.
First, the per-API span graph separates heterogeneous education workflows, so an exam submission path does not need to share the same representation as lightweight learning-content browsing.
Second, the trace-context readout and temporal decoder jointly model request intensity, recent latency percentiles, and failure-related signals, which are important when exam traffic rises quickly and then falls after the exam window.

The Edu Platform also exposes a useful limitation of the results: STLGT has the best MAPE but not the best MAE.
FastPERT reports a lower absolute error on this dataset, suggesting that it can fit the dominant low-latency operating range well.
However, education operators often care about relative degradation during high-pressure intervals, because a similar absolute error can correspond to very different operational risks when latency rapidly moves from routine learning traffic to exam traffic.
From this perspective, STLGT's lower MAPE is more aligned with burst-sensitive early warning, while its slightly higher MAE suggests room for further calibration of the absolute latency level in low-latency periods.

Overall, the Edu Platform case supports the applicability of STLGT beyond public microservice benchmarks.
It is particularly suitable for education scenarios where predictable schedules create sudden traffic concentration and where proactive prediction must distinguish normal learning activity from exam-time pressure.

\subsubsection{Discussion: Generalization and Topology Drift}
\label{sec:exp-discussion}

STLGT is train\-ed per API trace stream and assumes that the dominant call path remains stable within an evaluation period. This design is motivated by the observation that, in large-scale microservice systems, SLO violations are typically dominated by a small number of critical request paths. Modeling each critical API trace stream separately allows the predictor to focus on these critical paths, while keeping retraining and adaptation costs manageable when changes occur.

Under this formulation, moderate runtime variations---such as service-instance scaling or workload fluctuations that preserve the dominant invocation structure---can be effectively captured through time-varying service metrics and trace-level context features. More importantly, representing each API as a span graph provides a stable and scalable abstraction that naturally partitions a large microservice cluster into per-trace subgraphs, allowing STLGT to remain applicable even as the overall system scales.

However, substantial topology drift---for example, persistent call-path switching, service insertion or removal, or routing-policy changes---may weaken the fixed span-graph assumption and degrade prediction accuracy. In such cases, the API-specific span graph should be re-induced from recent traces, and the predictor should be fine-tuned or retrained on the updated stream. From this perspective, online adaptation to topology drift remains a limitation of the current offline evaluation rather than a solved problem.

We note that alternative approaches, such as subgraph partitioning in SGFormer~\cite{sgformer}, aim to improve scalability by splitting large graphs into smaller components. However, our empirical results suggest that such hard partitioning can be unstable for tail-latency prediction, potentially due to heterogeneous feature distributions across request paths in high-dimensional spaces. In contrast, the per-trace modeling strategy adopted by STLGT avoids imposing artificial graph boundaries and instead enables a more faithful representation of request-specific dependency structures. Moreover, this per-trace design naturally supports system-level composition by combining predictions from multiple trace-specific models with their corresponding topologies to provide more robust and reliable latency estimation.

%% file: sections/6-related-work.tex
\section{Related Work}
\label{sec:related-work}

% Microservice QoS forecasting and performance estimation are core components of cloud-native resource management, supporting proactive auto-scaling \cite{autopilot2020rzadca,aware}, SLO-oriented provisioning \cite{firm,autothrottle,zhang2024ursa}, and performance diagnosis and root cause analysis \cite{gan2019seer,xin2023causalrca,tao2024diagnosing,pham2024baro,pham2025rcaeval}. Existing studies can be broadly categorized into three classes: model-based approaches, learning-based approaches, and graph-based approaches \cite{surveyautoscaling2018chen,serveycloud2024deng}.

Microservice QoS prediction and performance estimation are fundamental to cloud-native resource management, enabling proactive auto-scaling \cite{autopilot2020rzadca,aware,firm,autothrottle,zhang2024ursa}, root cause analysis \cite{tao2024diagnosing,xin2023causalrca,pham2025rcaeval,pham2024baro}, and performance diagnosis \cite{gan2019seer,kadlecová2024pp}. 
Among various QoS metrics, tail latency at high percentiles such as p95 and p99 is particularly critical yet difficult to predict in complex microservice systems.
Existing studies can be broadly categorized into model-based, learning-based, and graph-based approaches \cite{surveyautoscaling2018chen,serveycloud2024deng}.

\subsection{Model-Based Approaches}
Model-based approaches characterize microservice performance using explicit analytical formulations, typically grounded in queueing theory or optimization models \cite{atom,mirhosseini2021parslo,rhythm}. Their primary advantages lie in low inference overhead and interpretability, making them suitable for capacity planning and scenarios with limited training data. Representative systems include ATOM \cite{atom} and several studies that allocate resources or decompose SLO budgets through analytical abstractions \cite{mirhosseini2021parslo,baarzi2021showar,rhythm}.

However, the effectiveness of model-based approaches critically depends on static modeling assumptions and accurate calibration. In modern production environments, frequent deployment change\-s, shared-resource interference, and heterogeneous execution behaviors often violate these assumptions, leading to large estimation errors, particularly for long dependency chains and tail-latenc\-y-sensitive workloads \cite{firm,pert}.

\subsection{Learning-Based Approaches}
Learning-based approaches leverage historical telemetry, such as metrics, logs, and traces, to train predictors for end-to-end QoS and integrate them into closed-loop resource management systems \cite{sinan,aware,firm}. Prior work has demonstrated their effectiveness in production settings, including QoS-aware auto-scaling \cite{sinan}, online scaling and SLO mitigation \cite{aware,firm}, bottleneck-aware resource adjustment \cite{pbscaler}, and fast recovery under dynamic workloads \cite{nodens,wang2024deepscaling}.

Despite their empirical success, learning-based methods face challenges under non-stationary workloads and configuration drift, often requiring frequent retraining and careful operational tuning. Moreover, many predictors rely on coarse-grained service-level aggregates or assume weak dependency structures, limiting their ability to capture cross-service influence propagation, long-chain error accumulation, and heterogeneous API-level invocation behaviors.

\subsection{Graph-Based Approaches}
To explicitly model service dependencies, topology-aware appro\-aches represent microservice systems as graphs, such as call graphs or trace-derived dependency graphs, and apply graph learning to encode interaction patterns \cite{deepscaler,graf,pert,fast-pert}. DeepScaler \cite{deepscaler} employs spatiotemporal GNNs with adaptive graph learning for sys\-tem-wide auto-scaling, while GRAF \cite{graf,park2024graph} learns graph representations to support SLO-oriented proactive resource allocation. PERT-GNN \cite{pert} models latency propagation using Trace-PERT gr\-aphs, and related work explores graph-based response-time prediction under different dependency abstractions \cite{graph-phpa,lq-gnn}.

While graph-based approaches improve dependency awareness, their modeling power often comes at the cost of scalability. Message passing or global interaction mechanisms can incur substantial training and inference overhead, and tightly coupled spatiotemporal designs further exacerbate complexity as system scale and workload dynamics increase \cite{deepscaler,graf,pert,lq-gnn}. In contrast, our work targets scalable, trace-driven QoS prediction by capturing global dependency propagation efficiently while decoupling temporal dynamics modeling to handle non-stationary and bursty workloads at production scale.

%% file: sections/7-conclusion.tex
\section{Conclusion}
\label{sec:conclusion}

Microservice tail-latency forecasting can enable proactive auto-sc\-aling, but practical predictors must address two competing challenges simultaneously: achieving high accuracy under non-station\-ary workloads and evolving dependency patterns, while remaining efficient enough for frequent inference at cluster scale.

We proposed STLGT, a scalable trace-based predictor for multi-step microservice tail-latency prediction.
STLGT constructs API-induced span graphs from distributed traces, aligns trace-derived features with service-level metrics, and learns to predict p95 end-to-end latency in future windows.
To capture long-range dependency propagation efficiently, STLGT adopts a linear-complexity graph Transformer encoder that combines structure-aware global mixing with lightweight local propagation, while a trace-aware readout and a decoupled temporal decoder model non-stationary workload dynamics without coupled spatiotemporal attention.
Experiments on microservice benchmarks~\cite{deathstarbench}, large-scale production traces~\cite{ali-traces}, and a personalized education platform show that STLGT improves forecasting accuracy over prior trace-graph baselines, while keeping inference overhead manageable as graph size increases.
The education case further indicates that STLGT is applicable to schedule-driven burst scenarios such as online exams, where accurate p95 prediction is needed before reactive scaling can absorb the traffic surge.

Our evaluation suggests that distributed traces contain informative signals for tail-latency forecasting, and that incorporating global context in the span-graph encoder can reduce prediction error.
We further find that the proposed linearized, structure-aware design provides a lightweight way to incorporate such global context, as supported by ablation results.
In future work, we plan to integrate STLGT into a closed-loop serving stack and evaluate end-to-end impact under production-grade auto-scaling controllers.

%% file: sections/8-ack.tex
\begin{acks}
This work was supported by the
\grantsponsor{nkrdp}{National Key Research and Development Program of China}{}
under Grant No.~\grantnum{nkrdp}{2023YFC3341200}.
\end{acks}

% 教育大数据驱动的个性化学习关键技术研究与示范应用（2023YFC3341200），国家重点研发计划 NKRDP